\documentclass[conference]{IEEEtran}
\IEEEoverridecommandlockouts

\usepackage{cite}
\usepackage{amsmath,amssymb,amsfonts}
\usepackage{algorithmic}
\usepackage{graphicx}
\usepackage{textcomp}
\usepackage{xcolor}
\usepackage{hyperref}
\usepackage{multirow}
\begin{document}

\title{Enhanced Multimodal RAG-LLM for Accurate Visual Question Answering}

\author{\IEEEauthorblockN{Junxiao Xue$^{1,*}$, Quan Deng$^{2,*}$, Fei Yu$^{1,\ddag}$, Yanhao Wang$^{3,\ddag}$, Jun Wang$^{1}$, Yuehua Li$^{1}$}
\IEEEauthorblockA{
$^{1}$\textit{Zhejiang Lab}\qquad$^{2}$\textit{Hangzhou Institute for Advanced Study, UCAS}\qquad $^{3}$\textit{East China Normal University}\\
\texttt{\{xuejx,yufei,wangjun,liyh\}@zhejianglab.org; dengquan23@mails.ucas.ac.cn;}\\
\texttt{yhwang@dase.ecnu.edu.cn}
}
\thanks{$^*$Both authors contributed equally to this work.}
\thanks{$^{\ddag}$Corresponding authors: Fei Yu and Yanhao Wang.}
}

\maketitle

\begin{abstract}
Multimodal large language models (MLLMs), such as GPT-4o, Gemini, LLaVA, and Flamingo, have made significant progress in integrating visual and textual modalities, excelling in tasks like visual question answering (VQA), image captioning, and content retrieval.
They can generate coherent and contextually relevant descriptions of images.
However, they still face challenges in accurately identifying and counting objects and determining their spatial locations, particularly in complex scenes with overlapping or small objects. To address these limitations, we propose a novel framework based on multimodal retrieval-augmented generation (RAG), which introduces structured scene graphs to enhance object recognition, relationship identification, and spatial understanding within images. Our framework improves the MLLM's capacity to handle tasks requiring precise visual descriptions, especially in scenarios with challenging perspectives, such as aerial views or scenes with dense object arrangements.
Finally, we conduct extensive experiments on the VG-150 dataset that focuses on first-person visual understanding and the AUG dataset that involves aerial imagery.
The results show that our approach consistently outperforms existing MLLMs in VQA tasks, which stands out in recognizing, localizing, and quantifying objects in different spatial contexts and provides more accurate visual descriptions.
\end{abstract}
    

\section{Introduction}
With recent advances in artificial intelligence techniques, multimodal large language models (MLLMs) have become central to integrating visual, textual, and other data modalities. Existing MLLMs, such as GPT-4 \cite{gpt4}, Gemini \cite{team2023gemini}, LLaVA \cite{NEURIPS2023_6dcf277e}, Flamingo \cite{alayrac2022flamingo}, PaLM-E \cite{driess2023palm}, MiniGPT-4 \cite{zhu2023minigpt}, along with earlier vision-language models like CLIP \cite{clip}, BLIP-2 \cite{blip}, and DALL-E \cite{dall-e}, have achieved impressive performance in tasks such as visual question answering (VQA) \cite{shao2023prompting}, image captioning \cite{NEURIPS2023_804b5e30}, and content retrieval \cite{gourav-etal-2024-multi}. These models are capable of recognizing objects and generating coherent and contextually relevant descriptions for various images.

\begin{figure}[t]
    \centering
    \includegraphics[width=\linewidth]{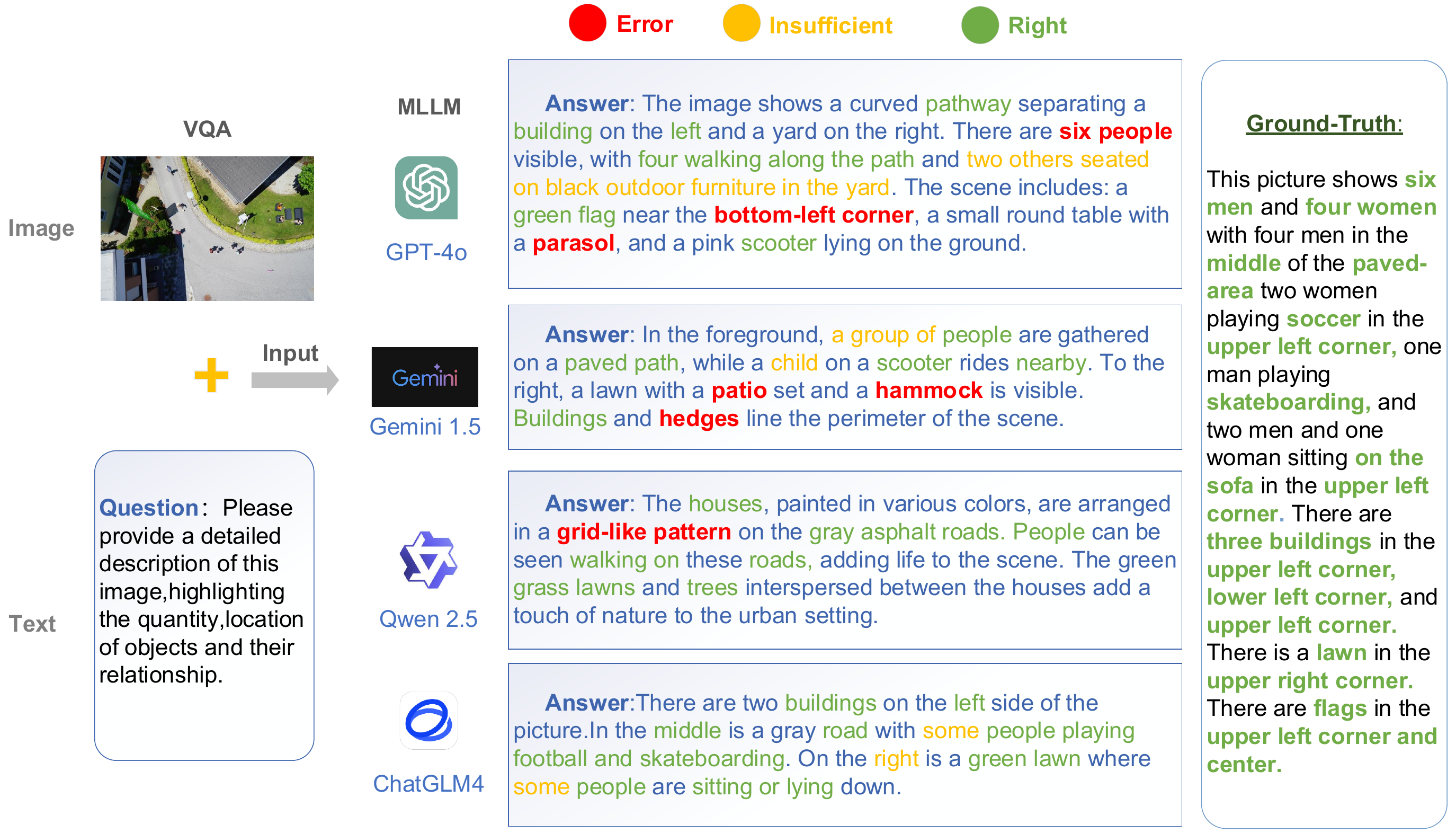}
    \caption{Examples of four mainstream MLLMs for VQA.}
    \label{fig:SamplePictures}
\end{figure}

However, despite their success in general vision-language tasks, these MLLMs face significant challenges in capturing complex visual details~\cite{mllm-duibi}. They often struggle with tasks like recognizing small and/or overlapped objects and accurately counting them, and identifying their spatial relationships, particularly in VQA tasks, as illustrated in Fig.~\ref{fig:SamplePictures}. Specifically, mainstream MLLMs such as GPT-4~\cite{gpt4}, Gemini~\cite{gemini}, Qwen~\cite{qwen-vl}, and ChatGLM~\cite{chatglm} frequently misidentify the quantity, location, and spatial relationships of different objects within an image, even if they have recognized most of the objects (e.g., people, buildings, and pathways). Moreover, these models lack sufficient domain-specific knowledge, limiting their ability to handle specialized images such as medical scans and technical diagrams~\cite{wu2024v, park2024assessing}. Their dependence on standardized datasets further exacerbates this problem, leading to suboptimal performance in complex VQA tasks~\cite{zang2024contextual, li2024survey}.

There have been several attempts to optimize MLLMs for VQA tasks.
These MLLM-based VQA methods focus mainly on improving object localization, which involves identifying the location, quantity, and orientation of an object within an image.
Although they have shown moderate performance \cite{chen2024gmai, huang2024survey}, there is considerable room for improvement in localization accuracy.
Furthermore, they focus mainly on the relative spatial relationships between objects (e.g., ``A above B'' and ``C next to D'') rather than the absolute locations of objects (e.g., ``A in the bottom right'' and ``B in the center'').
Additionally, existing methods for object quantity recognition often remain limited to total counts of all objects but cannot provide more fine-grained per-category counts (e.g., ``people'': 10 and ``houses'': 5).
All of these issues indicate the potential for further enhancement in localization and counting.

In (M)LLMs, hallucination is another critical issue to address.
In particular, hallucination often arises from the limitation of training data, such as outdated or insufficient domain-specific knowledge \cite{unreliability, unreliability2}.
Retrieval-Augmented Generation (RAG) can reduce hallucination by incorporating external knowledge into MLLMs.
RAG retrieves the relevant information from a knowledge database, typically of vectors, and uses it to constrain and enhance the model output.
When applying RAG to VQA tasks, for (single-modal) LLMs that can only process textual data, it is necessary to introduce a visual module to extract information from images.
For example, the optical character recognition (OCR) technique can extract text from images \cite{muti-RAG}, and an object detector such as YOLO-v8 can identify disease locations and types in plants from images \cite{kumar2024overcoming}.
Similarly, to allow an LLM to accurately answer questions related to object quantity, location, and relationships, a structured scene graph (SSG) generation module \cite{penet} should be added to the RAG framework to extract such information from images.

\begin{figure*}[t]
    \centering
    \includegraphics[width=\linewidth]{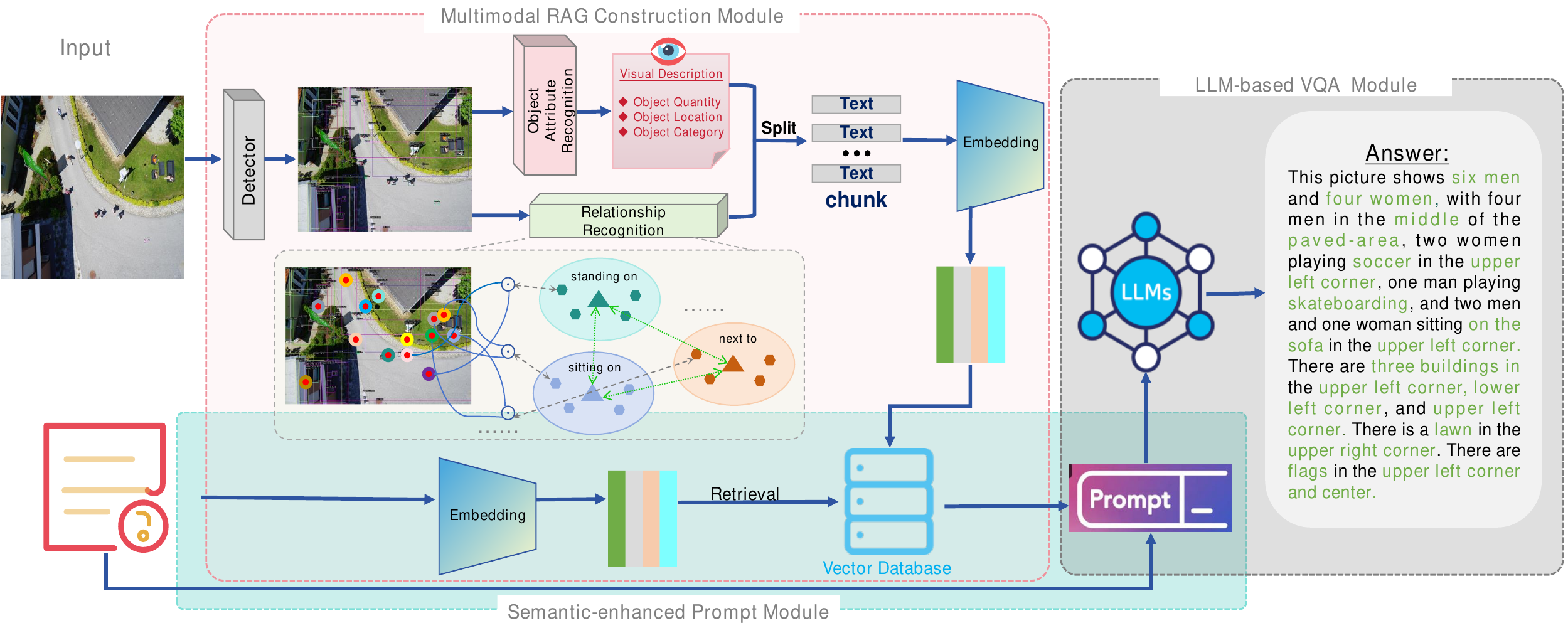}
    \caption{An overview of the enhanced multimodal RAG-LLM framework for accurate VQA.}
    \label{fig:framework}
\end{figure*}

Generally, existing MLLMs still fail to accurately handle VQA tasks involving object quantity, location, and spatial relationships.
In particular, they struggle to capture scenes accurately from complex and domain-specific visual contexts.
Existing MLLM-based methods for object localization and recognition cannot yet provide precise absolute locations and category-specific counts of objects.
None of the existing methods has considered incorporating structured scene graphs into MLLM-based VQA.
These limitations highlight the need for an improved framework for integrating and using multimodal information in VQA tasks.

To address the above challenges, we propose a new multimodal retrieval augmented generation (RAG) framework integrated with LLMs to improve visual reasoning in VQA tasks \cite{fan2024survey,jeong2023generative}.
Our framework constructs a domain-specific vector database by introducing structured scene graphs and improving object recognition and relationship identification.
Then, by dynamically retrieving and aligning relevant data with the input images and questions, our framework enhances an LLM's capability for accurate visual reasoning.
It significantly improves the accuracy of visual descriptions and complex object relationships by leveraging both the external knowledge of RAG and the general language capability of LLMs.
The key contributions of this paper are summarized as follows.
\begin{itemize} 
    \item We introduce an enhanced multimodal RAG-LLM framework that integrates multimodal data to improve the accuracy of visual descriptions in LLMs, addressing the deficiencies of existing MLLMs in small object recognition, exact counting, spatial positioning, and relationship identification. 
    \item We employ multimodal RAG for efficient extraction and processing of image structures without persistent storage, enabling accurate answers in VQA tasks by aligning image information with query context.
    \item We conduct extensive experiments to compare our model with existing MLLMs on two publicly available datasets (i.e., VG-150 and AUG) that focus on first-person visual understanding and involve aerial imagery, respectively. The results demonstrate the superior accuracy of our method in VQA tasks, especially in scenarios that require fine-grained object quantity, location, and relationships.
\end{itemize}

\section{Methodology}

In this section, we propose an enhanced multimodal RAG-LLM framework for accurate VQA tasks.
Fig.~\ref{fig:framework} illustrates an overview of our framework.
It consists of three key modules: (1) \emph{multimodal RAG construction}, which integrates visual and textual information to generate enriched representations; (2) \emph{semantic-enhanced prompt}, which further refines these representations by incorporating additional contextual information to generate the prompt; and (3) \emph{LLM-based VQA}, which performs the VQA task using an LLM with the generated prompt.
Next, we will describe each module in detail.

\subsection{Multimodal RAG Construction Module}

The primary goal of the multimodal RAG (Retrieval-Augmented Generation) construction module is to generate accurate answers to user queries by leveraging structured visual understanding of input images.
This involves generating a scene graph that identifies the entities (objects) in the image and their attributes (category, quantity, location, etc.) as well as the spatial relationships between them.

\smallskip
\noindent\textbf{Structured Visual Understanding:}
Given an input image $I$, the framework first generates a scene graph $G(I)$ that includes the objects in the image, the attributes of each object, and the spatial relationships between these objects, which are crucial for understanding the scene.
To generate the scene graph $G(I)$, we first adopt Faster-RCNN \cite{girshick2015fast} to produce entity proposals with corresponding features, which form the basis for object recognition.
Then, based on these proposals, we perform object attribute recognition to identify the key properties of each object, including its category, bounding box, quantity, and spatial location.
Specifically, the detector first identifies a set of objects $O = \{o_1, o_2, \ldots, o_n\}$ in the image $I$.
Then, for each detected object $o_i$, we assign several key attributes to it.
We first characterize $o_i$ by its type $t_i$ (e.g., ``person'', ``tree'', and ``building'').
Then, we use a bounding box $b_i$ to indicate the location and size of $o_i$ in the image $I$.
The bounding box $b_i$ is represented by the coordinates $(x_{\text{min}}, y_{\text{min}}, x_{\text{max}}, y_{\text{max}})$ of its four corners to locate $o_i$ in the image $I$.
Furthermore, for each object category $t_i$, we count the total number of its instances $N_{t_i}$ within the image $I$.
This count provides an overview of how many objects of each type are presented in the scene.
Finally, to determine the spatial location of each object $o_i$, we calculate the center point $c_i$ of its bounding box $b_i$, that is, 
\begin{equation*}
    c_i = \left(\frac{x_{\text{min}} + x_{\text{max}}}{2}, \frac{y_{\text{min}} + y_{\text{max}}}{2} \right).
\end{equation*}
We divide the image $I$ into a uniform grid of $3 \times 3$ regions (namely, `top', `center', and `bottom' vertically and `left', `center', `right' horizontally).
The center point $c_i$ of the object $o_i$ is mapped to one of the nine regions containing it (e.g., `top-left', `center', and `bottom-right') to describe its relative location within the scene of the image $I$.
By combining the category, the bounding box, the quantity, and the spatial location of each object, we generate a structured visual representation of the scene. This structured representation will serve as an input in subsequent procedures.

\smallskip
\noindent\textbf{Relationship Identification:}
To determine the relationships between different objects in the set $O$, we adopt the prototype-based embedding network (PENET) proposed in \cite{penet}.
PENET maps the object and predicate representations into a common semantic space to facilitate relationships recognition.
In particular, each relationship is presented by a triple $(o_i, p, o_j)$, where $o_i, o_j \in O$ are the subject and the object, and $p$ is the predicate.
We model each subject $o_i$, object $o_j$, and predicate $p$ using class-specific prototypes.
The representations of $o_i$, $o_j$ and $p$ are expressed by $\mathbf{o}_i = \mathbf{W}_{i} \mathbf{t}_i + \mathbf{v}_i$, $\mathbf{o}_j = \mathbf{W}_{j} \mathbf{t}_j + \mathbf{v}_j$, and $\mathbf{p} = \mathbf{W}_p \mathbf{t}_p + \mathbf{u}_p$, where $\mathbf{W}_{i}$, $\mathbf{W}_{j}$, and $\mathbf{W}_p$ are learnable parameters; $\mathbf{W}_{i} \mathbf{t}_i$, $\mathbf{W}_{j} \mathbf{t}_j$, and $\mathbf{W}_p \mathbf{t}_p$ are semantic prototypes obtained from word embeddings such as GloVe.
Based on PENET's class-specific prototypes, the vectors $\mathbf{v}_i$, $\mathbf{v}_j$, and $\mathbf{u}_p$ are used to model the instance-varied semantic contents in their respective classes.
Here, $\mathbf{v}_i$ and $\mathbf{u}_p$ are obtained as
\begin{align*}
    & \mathbf{v}_i = \sigma(\mathrm{FC}(\mathbf{W}_{i} \mathbf{t}_i, \mathcal{M}(\mathbf{e}_{i}))) \odot \mathcal{M}(\mathbf{e}_{i});\\
    & \mathbf{u}_p = \sigma(\mathrm{FC}(\mathcal{G}(\mathbf{o}_i, \mathbf{o}_j),\mathcal{M}(\mathbf{e}_p))) \odot \mathcal{M}(\mathbf{e}_p),
\end{align*}
where $\sigma(\cdot)$ is the activation function, $\mathrm{FC}(\cdot)$ is a fully connected layer, $\mathcal{M}(\cdot)$ represents a visual-to-semantic transformation that converts the visual features into the semantic space, $\mathbf{e}_{i}$ and $\mathbf{e}_p$ are the visual features of $o_i$ and the union of $o_i$ and $o_j$ from the object detection module, and $\mathcal{G}(\mathbf{o}_i, \mathbf{o}_j) = \mathrm{ReLU}(\mathbf{o}_i + \mathbf{o}_j) - (\mathbf{o}_i - \mathbf{o}_j)^2$ is the feature fusion function.
Moreover, $\mathbf{v}_j$ is obtained in the same way as $\mathbf{v}_i$ by replacing all $i$'s with $j$'s.
Based on the representations of objects and relationships by PENET, we construct a vector database as follows.

\smallskip
\noindent\textbf{Vector Database Construction:}
The vector database can enable efficient retrieval of relevant information from pre-computed vectors.
First, we organize the identified objects and relationships, along with their attributes such as quantity, location, and category, into coherent chunks.
Each chunk represents a specific object category, along with its quantity, location, and relationships.
An exemplar chunk may look like \textit{``car: 3, location: [center, left], relationships: car near tree, man in car''}.
This structured representation of information allows for streamlined retrieval when addressing user queries.
Then, each chunk is embedded into a high-dimensional vector space using pre-trained models.\footnote{In our implementation, Text2Vec Base Multilingual (Version 1.0) is used.}
The embedding process transforms the attributes and relationships of objects into dense vector representations, which are crucial for semantic comparison and retrieval.
These embedding vectors are then stored in a temporary vector database, where each vector is mapped to its corresponding chunk.
During the retrieval phase, the query vector is compared with the pre-computed vectors to identify the top-4 closest matches with the highest cosine similarities. The most relevant information, derived from these matches, is integrated with the query to generate contextually relevant responses.

\subsection{Semantic-enhanced Prompt Module}

In the semantic-enhanced prompt generation module, the framework improves its understanding and response generation by integrating the retrieved visual data with the user's question in a structured format.
After retrieving the top-4 relevant chunks of information (e.g., object counts, locations, and relationships) from the vector database, the information is combined into a prompt to semantically enhance the LLM's ability to process the query.
Specifically, the prompt is composed as ``\textit{Based on the information extracted from the image: \{\texttt{DATA}\}, please answer the following question: \{\texttt{QUESTION}\}}.''

In the above process, semantic enhancement comes from the data component that provides the LLM with contextually relevant visual information about the image.
This ensures that the LLM is not only working from the query, but also has access to detailed, structured image data that have been optimized through semantic retrieval.

\subsection{LLM-based VQA Module}

In the LLM-based VQA module, the semantic-enhanced prompt is fed to the LLM to produce more accurate and context-aware answers. Here, the LLM we use in the implementation is Qwen-2-72B-Instruct.\footnote{Note that any other LLM can also work in our framework, as long as it can understand our prompts.}
The retrieved answers encapsulate important object-level and relational details from the image, which, when combined with the query, allows the LLM to operate in a semantically rich environment.
This enhancement effectively mitigates the ambiguity that might arise if the LLM is only processing the query, leading to more precise and context-sensitive responses.
The semantic enhancement process is crucial for processing VQA tasks in complex scenarios, as it allows the LLM to base its reasoning not only on linguistic information but also on a structured and semantically aligned understanding of the visual content. This results in a more informed and contextually aware generation of answers.

\section{Experiments}

\subsection{Experimental Setup}

\noindent\textbf{Datasets:}
We used two real-world datasets in the experiments.
The visual genome (VG) dataset \cite{vg} consists of 108,077 images, most of which are in first person, with an average of 38 objects and 22 relationships per image.
Due to the sparsity of certain object categories, we used the VG-150 subset \cite{vg150} with the 150 most frequent object categories and 50 predicates between them.
We sampled 100 images from the VG-150 test set to evaluate each method.
The AUG dataset \cite{aug} consists of 400 images of aerial urban views with smaller, more densely packed objects, encompassing 77 object categories and 63 relationship types, with an average of 63 objects and 42 relationships per image.
For AUG, we used 300 images for training and the remaining 100 for testing.

\smallskip
\noindent\textbf{Evaluation Metrics:}
To evaluate the performance of each model on VQA, we use the following metrics.
First, the \emph{recall} scores are used to measure the ratios of ground-truth objects, quantities, locations, and relationships that are correctly described in the answer for an image returned by a specific model.
Then, we also use the \emph{F1-scores}, which combines \emph{recall} with \emph{precision} to further reflect the ratios of incorrect objects, quantities, locations, and relationships in the answer of a specific model.
Finally, we calculate an \emph{overall score} for each model as the ratio of classes (on all four attributes) in which the model achieves recall scores of at least 0.55.
This can reflect the overall performance of each model in various scenarios.

\noindent\textbf{Implementation:}
The model is deployed on a server with an Intel Core i7-11700 CPU, 32GB RAM, and an NVIDIA GeForce RTX 3060 GPU with 12GB of dedicated memory.
The implementation leverages PyTorch version 2.3.1 along with CUDA 12.1.
To fairly compare the VQA ability of different methods, we used an integrated prompt to obtain the categories, quantities, locations, and relationships of various objects within each image from each MLLM.

\smallskip
\noindent\textbf{Baselines:}
We compare our proposed method with three mainstream MLLMs: (1) GPT-4o~\cite{gpt4} (ver.~20240806), which is top-ranked in the latest MLLM leaderboard~\cite{opencompass}; (2) Qwen-VL-Max~\cite{qwen-vl}; and (3) Gemini Flash~\cite{gemini} (v1.5), both of which are popular MLLMs that have also shown strong capacities in VQA tasks.

\subsection{Experimental Results}

\begin{table}[t]
    \centering
    \setlength{\tabcolsep}{3pt}
    \caption{The recall scores of different methods on the VG-150 and AUG test sets. The bold font indicates the best performance among all compared methods for each attribute.}
    \label{tab:VG150_AUG_recall}
    \resizebox{\linewidth}{!}{
    \begin{tabular}{|c|c|cccc|}
        \hline
        \textbf{Dataset} & \textbf{Model} & \textbf{Category} & \textbf{Quantity} & \textbf{Location} & \textbf{Relationship} \\
        \hline
        \multirow{4}{*}{VG-150} & GPT-4o       & 0.2586 & 0.1475 & 0.0719 & 0.0295 \\
                                & Qwen-VL-Max  & 0.3017 & 0.2032 & 0.0764 & 0.0505 \\
                                & Gemini Flash & 0.2304 & 0.1355 & 0.0630 & 0.0327 \\
                                & Ours (*)     & \textbf{0.5503} & \textbf{0.3468} & \textbf{0.1308} & \textbf{0.1172} \\
        \hline
        \multirow{4}{*}{AUG}    & GPT-4o       & 0.0908 & 0.0584 & 0.0324 & 0.0001 \\
                                & Qwen-VL-Max  & 0.0877 & 0.0508 & 0.0229 & 0.0055 \\
                                & Gemini Flash & 0.1615 & 0.0908 & 0.0555 & 0.0095 \\
                                & Ours (*)     & \textbf{0.6494} & \textbf{0.3119} & \textbf{0.3311} & \textbf{0.1684} \\
        \hline
    \end{tabular}
    }
\end{table}

As shown in Table~\ref{tab:VG150_AUG_recall}, our enhanced multimodal RAG-LLM method significantly outperforms three baseline MLLMs in terms of recall scores for VQA in the four attributes (object categories, quantities, locations, and relationships).
In the VG-150 dataset, it achieves improvements of 108.84\%, 113.94\%, 87.79\%, and 212.53\% on average in each attribute, respectively.
Although Qwen-VL-Max \cite{qwen-vl} has been trained on VG-150 and outperforms GPT-4o and Gemini Flash, our method still shows substantial improvements of 75.17\%, 70.67\%, 71.20\%, and 132.07\% compared to Qwen-VL-Max.
In the AUG dataset with smaller objects and more complex relationships, our method shows even greater improvements in recall scores.
Specifically, we observe improvements of 473.17\%, 367.62\%, 797.28\%, and 3268.00\% on average in the four attributes.
Compared to Gemini Flash, the best performing MLLM among the three, our method improves the recall scores by 302.11\%, 243.50\%, 496.58\%, and 1672.63\%.
These results highlight the limitations of current MLLMs in small object detection, counting, spatial reasoning, and relationship identification, and confirm the efficacy of our method in these tasks.

\begin{table}[t]
    \centering
    \setlength{\tabcolsep}{3pt}
    \caption{The F1-scores of different methods on the VG-150 and AUG test set. The bold font indicates the best performance among all compared methods for each attribute.}
    \label{tab:VG150_AUG_F1}
    \resizebox{\linewidth}{!}{
    \begin{tabular}{|c|c|cccc|}
        \hline
        \textbf{Dataset} & \textbf{Model} & \textbf{Category} & \textbf{Quantity} & \textbf{Location} & \textbf{Relationship} \\
        \hline
        \multirow{4}{*}{VG-150} & GPT-4o       & 0.3427 & 0.1918 & 0.0886 & 0.0255 \\
                                & Qwen-VL-Max  & 0.3412 & 0.2257 & 0.0853 & 0.0377 \\
                                & Gemini Flash & 0.3000 & 0.1748 & 0.0662 & 0.0296 \\
                                & Ours (*)     & \textbf{0.4973} & \textbf{0.3056} & \textbf{0.1443} & \textbf{0.0459} \\
        \hline
        \multirow{4}{*}{AUG}    & Qwen-VL-Max  & 0.1387 & 0.0778 & 0.0345 & 0.0075 \\
                                & Gemini Flash & 0.2488 & 0.1389 & 0.0800 & 0.0142 \\
                                & GPT-4o       & 0.1472 & 0.0921 & 0.0504 & 0.0002 \\
                                & Ours (*)     & \textbf{0.7156} & \textbf{0.3350} & \textbf{0.3963} & \textbf{0.1184} \\
        \hline
    \end{tabular}
    }
\end{table}

As shown in Table~\ref{tab:VG150_AUG_F1}, our method also achieves the highest F1-scores in the VG-150 and AUG datasets. 
In the VG-150 dataset, our method achieves average improvements of 51.62\%, 54.81\%, 80.37\%, and 48.54\%.
In the AUG dataset, similar to those of the recall scores, the average increases in the F1-scores are more substantial, namely 301.57\%, 225.56\%, 336.54\%, and 1521.91\%.
These results indicate that our method not only achieves higher coverage but also makes fewer errors in VQA and that introducing structural information and external knowledge largely avoids hallucination in MLLMs.

\begin{table}[t]
    \centering
    \caption{The overall scores of different methods on the VG-150 and AUG test set. The bold font indicates the best performance on each dataset.}
    \label{tab:Overall score}
    \vspace{-1em}
    \begin{tabular}{|c|c|c|c|c|}
        \hline
        Dataset & Qwen-VL-Max & Gemini Flash & GPT-4o & Ours (*) \\
        \hline
        VG-150 & 0.0500 & 0.0250 & 0.0250 & \textbf{0.1950} \\
        AUG    & 0.0051 & 0.0000 & 0.0000 & \textbf{0.2250} \\
        \hline
    \end{tabular}
\end{table}
\begin{figure}[t]
    \centering
    \includegraphics[width=\linewidth]{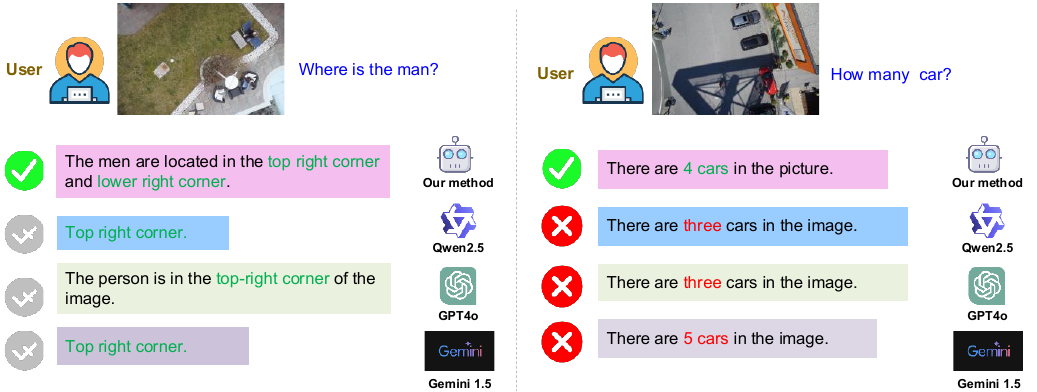}
    \vspace{-1em}
    \caption{Comparative analysis of our method and baseline MLLMs for VQA.}
    \label{fig:Pictures3}
\end{figure}

Table~\ref{tab:Overall score} presents the overall scores of different methods in the VG-150 and AUG datasets.
Due to the subpar performance of MLLMs in recognizing categories, quantities, locations, and relationships, as evidenced by Table~\ref{tab:VG150_AUG_recall}, they can achieve a recall score of at least 0.55 only for a few classes.
As such, their overall scores are very low. In the AUG dataset, Gemini Flash and GPT-4o cannot achieve a recall score of 0.55 in any single class, resulting in an overall score of $0$.
Our method significantly outperforms the baseline MLLMs in overall scores, particularly in the AUG dataset.

Fig.~\ref{fig:Pictures3} illustrates a comparative analysis of our method with baseline MLLMs in VQA tasks.
As demonstrated by examples, our method excels at understanding complex scenes and accurately counting and positioning objects within them, which are often challenging for baseline MLLMs.
Such capacities are very important for the deployment of MLLMs in many real-world applications, such as remote sensing, robotics, and the Internet of Things (IoT).

In general, our method consistently outperforms three baseline MLLMs across two datasets, mainly focusing on two views: first-person and aerial views.
In both datasets, our method achieves significantly higher recall scores and F1-scores, particularly excelling at recognizing, localizing, and quantifying objects in complex spatial contexts.
These results confirm our method's effectiveness in accurate VQA tasks, showcasing its adaptability across diverse perspectives and its advantages over baseline MLLMs.

\section{Conclusion}

In this paper, we proposed a novel multimodal framework that combines RAG and structured scene graphs to enhance VQA tasks.
Our approach significantly improved the accuracy of visual descriptions, achieving good performance in small object detection, exact counting, spatial localization, and relationship identification, where existing MLLMs often struggled.
Compared to state-of-the-art MLLMs, our method achieved better results in all four dimensions.
Furthermore, our method is adaptable and can easily incorporate domain-specific knowledge through RAG, which is effective for VQA and other complex multimodal applications.

\bibliographystyle{IEEEtran}
\bibliography{reference} 

\end{document}